\documentclass[sigconf, nonacm]{acmart}
\usepackage[ruled,vlined]{algorithm2e}
\usepackage{amsfonts}
\usepackage{amsmath}
\usepackage{pgf}

\renewcommand{\vec}[1]{\mathbf{#1}}
\DeclareMathAlphabet{\mathcal}{OMS}{cmsy}{m}{n}

\AtBeginDocument{%
  \providecommand\BibTeX{{%
    \normalfont B\kern-0.5em{\scshape i\kern-0.25em b}\kern-0.8em\TeX}}}

\setcopyright{rightsretained}
\copyrightyear{2020}
\acmYear{2020}
\acmDOI{10.1145/1122445.1122456}

\acmConference[KDD '20]{KDD '20}{August 22--27, 2020}{San Diego, CA}
\acmBooktitle{KDD '20, August 22--27, 2020, San Diego, CA}
\acmISBN{978-1-4503-XXXX-X/18/06}

\begin{document}

\title{Out-of-Core GPU Gradient Boosting}

\author{Rong Ou}
\affiliation{%
  \institution{NVIDIA}
  \city{Santa Clara}
  \state{CA}
  \country{USA}
}
\email{rou@nvidia.com}

\begin{abstract}
GPU-based algorithms have greatly accelerated many machine learning methods;
however, GPU memory is typically smaller than main memory, limiting the size of training data.
In this paper, we describe an out-of-core GPU gradient boosting algorithm implemented in the
XGBoost library.
We show that much larger datasets can fit on a given GPU, without degrading model accuracy or
training time.
To the best of our knowledge, this is the first out-of-core GPU implementation of gradient
boosting.
Similar approaches can be applied to other machine learning algorithms.
\end{abstract}



\maketitle

\section{Introduction}
\label{sec:introduction}
Gradient boosting~\cite{Friedman2001} is a popular machine learning method for supervised
learning tasks, such as classification, regression, and ranking.
A prediction model is built sequentially out of an ensemble of weak prediction models, typically
decision trees.
With bigger datasets and deeper trees, training time can become substantial.

Graphics Processing Units (GPUs), originally designed to speed up the rendering of display images,
have proven to be powerful accelerators for many parallel computing tasks, including machine
learning.
GPU-based implementations~\cite{XGBoost2020GPU, LightGBM2020, CatBoost2020} exist for several
open-source gradient boosting libraries~\cite{Chen2016, Ke2017, Prokhorenkova2018} that
significantly lower the training time.

Because GPU memory has higher bandwidth and lower latency, it tends to cost more and thus is
typically of smaller size than main memory.
For example, on Amazon Web Services (AWS), a \verb|p3.2xlarge| instance has 1 NVIDIA Tesla V100
GPU with 16 GiB memory, and 61 GiB main memory.
On Google Cloud Platform (GCP), a similar instance can have as much as 78 GiB main memory.
Training with large datasets can cause GPU out-of-memory errors when there is plenty of main
memory available.

XGBoost, a widely-used gradient boosting library, has experimental support for external
memory~\cite{XGBoost2020ExternalMemory}, which allows training on datasets that do not fit in
main memory \footnote{In this paper, "out-of-core" and "external memory" are used
interchangeably.}.
Building on top of this feature, we designed and implemented out-of-core GPU algorithms that
extend XGBoost external memory support to GPUs.
This is challenging since GPUs are typically connected to the rest of the computer system through
the PCI Express (PCIe) bus, which has lower bandwidth and higher latency than the main memory bus.
A naive approach that constantly swaps data in and out of GPU memory would cause too much
slowdown, negating the performance gain from GPUs.

By carefully structuring the data access patterns, and leveraging gradient-based sampling to
reduce working memory size, we were able to significantly increase the size of training data
accommodated by a given GPU, with minimal impact to model accuracy and training time.

\section{Background}
\label{sec:background}
In this section we review the gradient boosting algorithm as implemented by XGBoost, its GPU
variant, and the previous CPU-only external memory support.
We also describe the sampling approaches used to reduce memory footprint.

\subsection{Gradient Boosting}
\label{subsec:gradient-boosting}
Given a dataset with $n$ samples $\{\vec{x_i}, y_i\}_{i=1}^{n}$, where
$\vec{x_i}\in{\mathbb{R}^{m}}$ is a vector of $m$-dimensional input features, and
$y_i\in{\mathbb{R}}$ is the label, a decision tree model predicts the label:
\begin{equation}
  \hat{y}_i = F(\vec{x_i}) = \sum_{k=1}^{K}{f_k(\vec{x_i})},
\end{equation}
where $f_k\in{\mathcal{F}}$, the space of regression trees, and $K$ is the number of trees.
To learn a model, we minimize the following regularized objective:
\begin{gather}
  \mathcal{L}(F) = \sum_{i}{l(\hat{y_i}, y_i)} + \sum_{k}\Omega(f_k) \\
  \text{where} \ \Omega(f) = \gamma T + \frac{1}{2} \lambda ||w||^2
\end{gather}
Here $l$ is a differentiable loss function, $\Omega$ is the regularization term that penalizes
the number of leaves in the tree $T$ and leaf weights $w$, controlled by two hyperparameters
$\gamma$ and $\lambda$.

The model is trained sequentially.
Let $\hat{y}_{i}^{(t)}$ be the prediction at the $t$-th iteration, we need to find tree $f_t$
that minimizes:
\begin{equation}
  \mathcal{L}^{(t)} = \sum_{i=1}^n{l(y_i , \hat{y}_i^{(t-1)} + f_t(\vec{x_i}))} + \Omega(f_t)
\end{equation}
The quadratic Taylor expansion is:
\begin{equation}
  \mathcal{L}^{(t)} \simeq \sum_{i=1}^n{[l(y_i , \hat{y}_i^{(t-1)}) + g_i f_t(\vec{x_i}) +
    \frac{1}{2} h_i f_t^2(\vec{x_i})]} + \Omega (f_t),
\end{equation}
where $g_i$ and $h_i$ are first and second order gradients on the loss function with respect to
$\hat{y}^{(t-1)}$.
For a given tree structure $q(\vec{x})$, let $I_j = \{i|q(\vec{x}_i) = j\}$ be the set of samples
that fall into leaf $j$.
The optimal weight $w_j^{*}$ of leaf $j$ can be computed as:
\begin{equation}
  w_j^{*} = - \frac{\sum_{i\in{I_j}} g_i}{\sum_{i\in{I_j}} h_i + \lambda},
\end{equation}
and the corresponding optimal loss value is:
\begin{equation}
  \tilde{\mathcal{L}}^{(t)}(q) = - \frac{1}{2} \sum_{j=1}^T \frac{(\sum_{i\in{I_j}} g_i)
    ^2}{\sum_{i\in{I_j}} h_i + \lambda} + \gamma T.
\end{equation}
When constructing an individual tree, we start from a single leaf and greedily add branches to
the tree.
Let $I_L$ and $I_R$ be the sets of samples that fall into the left and right nodes after a split,
then the loss reduction for a split is:
\begin{equation}
  \mathcal{L}_{split} = \frac{1}{2} \Bigg[
    \frac{(\sum_{i\in{I_L}} g_i)^2}{\sum_{i\in{I_L}} h_i + \lambda} +
    \frac{(\sum_{i\in{I_R}} g_i)^2}{\sum_{i\in{I_R}} h_i + \lambda} -
    \frac{(\sum_{i\in{I}} g_i)^2}{\sum_{i\in{I}} h_i + \lambda}
  \Bigg] - \gamma
\end{equation}
where $I = I_L \cup I_R$.

\subsection{GPU Tree Construction}
\label{subsec:gpu-tree-construction}
The GPU tree construction algorithm in XGBoost~\cite{Mitchell2017, Mitchell2018} relies on a
two-step process.
First, in a preprocessing step, each input feature is divided into quantiles and put into bins
(\verb|max_bin| defaults to 256).
The bin numbers are then compressed into \verb|ELLPACK| format, greatly reducing the size of the
training data.
This step is time consuming, so it should only be done once at the beginning of training.

\begin{algorithm}
  \KwIn{$X$: training examples}
  \KwIn{$g$: gradient pairs for training examples}
  \KwOut{$tree$: set of output nodes}
  tree $\leftarrow$ \{ \}\\
  queue $\leftarrow$ InitRoot()\\
  \While{queue is not empty}{
    entry $\leftarrow$ queue.pop()\\
    tree.insert(entry)\\
    \tcp{Sort samples into leaf nodes}
    RepartitionInstances(entry, $X$)\\
    \tcp{Build gradient histograms}
    BuildHistograms(entry, $X$, $g$)\\
    \tcp{Find the optimal split for children}
    left\_entry $\leftarrow$ EvaluateSplit(entry.left\_histogram)\\
    right\_entry $\leftarrow$ EvaluateSplit(entry.right\_histogram)\\
    queue.push(left\_entry)\\
    queue.push(right\_entry)\\
  }
  \caption{GPU Tree Construction}
  \label{alg:gpu-tree-construction}
\end{algorithm}

In the second step, the tree construction algorithm is shown in
Algorithm~\ref{alg:gpu-tree-construction}.
Note that this is a simplified version for single GPU only.
In a distributed environment with multiple GPUs, the gradient histograms need to be summed across
all GPUs using \verb|AllReduce|.

\subsection{XGBoost Out-of-Core Computation}
\label{subsec:xgboost-out-of-core-computation}
XGBoost has experimental support for out-of-core computation~\cite{Chen2016,
XGBoost2020ExternalMemory}.
When enabled, training is also done in a two-step process.
First, in the preprocessing step, input data is read and parsed into an internal format,
which can be Compressed Sparse Row (CSR), Compressed Sparse Column (CSC), or sorted CSC.
Each sample is appended to an in-memory buffer.
When the buffer reaches a pre-defined size (32 MiB), it is written out to disk as a \verb|page|.
Second, during tree construction, the data pages are streamed from disk via a multi-threaded
pre-fetcher.

\subsection{Sampling}
\label{subsec:sampling}
In its default setting, gradient boosting is a batch algorithm: the whole dataset needs to be
read and processed to construct each tree.
Different sampling approaches have been proposed, mainly as an additional regularization factor
to get better generalization performance, but they can also reduce the computation needed,
leading to faster training time.

\subsubsection{Stochastic Gradient Boosting (SGB)}
Shortly after introducing gradient boosting, Friedman~\cite{Friedman2002} proposed an
improvement: at each iteration a subsample of the training data is drawn at random without
replacement from the full training dataset.
This randomly selected subsample is then used in place of the full sample to construct the decision
tree and compute the model update for the current iteration.
It was shown that this sampling approach improves model accuracy.
However, the sampling ratio, $f$, needs to stay relatively high, $0.5 \leq f \leq 0.8$, for this
improvement to occur.

\subsubsection{Gradient-based One-Side Sampling (GOSS)}
Ke \textit{et al}.
proposed a sampling strategy weighted by the absolute value of the gradients~\cite{Ke2017}.
At the beginning of each iteration, the top $a \times 100\%$ of training instances with the
largest gradients are selected, then from the rest of the data a random sample of $b \times
100\%$ instances is drawn.
The samples are scaled by $\frac{1-a}{b}$ to make the gradient statistics unbiased.
Compared to SGB, GOSS can sample more aggressively, only using 10\% - 20\% of the data to achieve
similar model accuracy.

\subsubsection{Minimal Variance Sampling (MVS)}
Ibragimov \textit{et al}.
proposed another gradient-based sampling approach that aims to minimize the variance of the model.
At each iteration the whole dataset is sampled with probability proportional to
\textit{regularized absolute value} of gradients:
\begin{equation}
  \hat{g}_i = \sqrt{g_i^2 + \lambda h_i^2},
\end{equation}
where $g_i$ and $h_i$ are the first and second order gradients, $\lambda$ can be either a
hyperparameter, or estimated from the squared mean of the initial leaf value.

MVS was shown to perform better than both SGB and GOSS, with sampling rate as low as 10\%.

\section{Method}
\label{sec:method}
In this section we describe the design of out-of-core GPU-based gradient boosting.
Since XGBoost is widely used in production, as much as possible, we try to preserve the existing
behavior when adding new features.
In external memory mode, we assume the training data is already parsed and written to disk in
CSR pages.

\subsection{Incremental Quantile Generation}
\label{subsec:incremental-quantile-generation}

\begin{algorithm}
  \KwIn{$X$: training examples}
  \KwOut{$histogram\_cuts$: cut points for all features}
  \ForEach{batch in X (a single CSR page)}{
    CopyToGPU(batch)\\
    \ForEach{column in batch}{
      cuts $\leftarrow$ FindColumnCuts(batch, column)\\
      CopyColumnCuts(histogram\_cuts, cuts)\\
    }
  }
  \caption{In-Core Quantile Sketch}
  \label{alg:in-core-quantile-sketch}
\end{algorithm}

As stated above, GPU tree construction in XGBoost is a two-step process.
In the preprocessing step, input features are converted into a quantile representation.
Quantiles are cut points dividing the range of each feature into continuous intervals
(\textit{i.e.} bins) with equal probabilities.
Algorithm~\ref{alg:in-core-quantile-sketch} shows the in-core version of quantile sketch.

\begin{algorithm}
  \KwIn{$X$: training examples}
  \KwOut{$histogram\_cuts$: cut points for all features}
  \ForEach{page in X}{
    \ForEach{batch in page}{
      CopyToGPU(batch)\\
      \ForEach{column in batch}{
        cuts $\leftarrow$ FindColumnCuts(batch, column)\\
        CopyColumnCuts(histogram\_cuts, cuts)\\
      }
    }
  }
  \caption{Out-of-Core Quantile Sketch}
  \label{alg:out-of-core-quantile-sketch}
\end{algorithm}

Since the existing code already operates in batches and handles the necessary bookkeeping, it is
straightforward to extend it to external memory mode with multiple CSR pages, as shown in
Algorithm~\ref{alg:out-of-core-quantile-sketch}.

\subsection{External ELLPACK Matrix}
\label{subsec:external-ellpack-matrix}

\begin{algorithm}
  \KwIn{$X$: training examples}
  \KwIn{$histogram\_cuts$: cut points for all features}
  \KwOut{$ellpack\_page$: compressed ELLPACK matrix}
  AllocateOnGPU(ellpack\_page)\\
  \ForEach{batch in X (a single CSR page)}{
    CopyToGPU(batch)\\
    \ForEach{row in batch}{
      \ForEach{column in row}{
        bin $\leftarrow$ LookupBin(histogram\_cuts, column)\\
        Write(ellpack\_page, bin)\\
      }
    }
  }
  \caption{In-Core ELLPACK Page}
  \label{alg:in-core-ellpack-page}
\end{algorithm}

Once the quantile cut points are found, input features can be converted to bin numbers and
compressed into ELLPACK format, as shown in Algorithm~\ref{alg:in-core-ellpack-page}.

\begin{algorithm}
  \KwIn{$X$: training examples}
  \KwOut{$ellpack\_pages$: compressed ELLPACK matrix pages}
  list $\leftarrow$ \{ \}\\
  \ForEach{page in X}{
    list.append(page)\\
    \If{CalculateEllpackPageSize(list) >= 32 MiB}{
      AllocateOnGPU(ellpack\_page)\\
      \ForEach{page in list}{
        Write(ellpack\_page, page)\\
      }
      WriteToDisk(ellpack\_page)\\
      list $\leftarrow$ \{ \}\\
    }
  }
  \tcp{Convert $list$ to ELLPACK and write to disk}
  \ldots\\
  \caption{Out-of-Core ELLPACK Pages}
  \label{alg:out-of-core-ellpack-pages}
\end{algorithm}

In external memory mode, we assume the single ELLPACK matrix may not fit in GPU memory, thus is
broken up into multiple ELLPACK pages and written to disk.
Since CSR pages contain variable number of rows, we cannot pre-allocate these ELLPACK pages.
Instead, the CSR pages are accumulated in memory first.
When the expected ELLPACK page reaches the size limit, the CSR pages are converted and written to
disk, as shown in Algorithm~\ref{alg:out-of-core-ellpack-pages}.

\subsection{Incremental Tree Construction}
\label{subsec:incremental-tree-construction}

\begin{algorithm}
  \KwIn{$X$: training examples}
  \KwIn{$g$: gradient pairs for training examples}
  \KwOut{$tree$: set of output nodes}
  tree $\leftarrow$ \{ \}\\
  \tcp{Loop through all the pages}
  queue $\leftarrow$ InitRoot()\\
  \While{queue is not empty}{
    entry $\leftarrow$ queue.pop()\\
    tree.insert(entry)\\
    \ForEach{page in X}{
      \tcp{Sort samples into leaf nodes}
      RepartitionInstances(entry, page)\\
      \tcp{Build gradient histograms}
      BuildHistograms(entry, page, $g$)\\
    }
    \tcp{Find the optimal split for children}
    left\_entry $\leftarrow$ EvaluateSplit(entry.left\_histogram)\\
    right\_entry $\leftarrow$ EvaluateSplit(entry.right\_histogram)\\
    queue.push(left\_entry)\\
    queue.push(right\_entry)\\
  }
  \caption{Naive Out-of-Core GPU Tree Construction}
  \label{alg:naive-out-of-core-tree-construction}
\end{algorithm}

Now we finally have the ELLPACK pages on disk, a naive tree construction method is to stream the
pages for each tree node, as shown in Algorithm~\ref{alg:naive-out-of-core-tree-construction}.
However, because of the PCIe bottleneck, this approach performed badly, even slower than the
CPU tree construction algorithm.

\subsection{Use Sampled Data}
\label{subsec:use-sampled-data}

\begin{algorithm}
  \KwIn{$X$: training examples}
  \KwIn{$g$: gradient pairs for training examples}
  \KwOut{$tree$: set of output nodes}
  $g' \leftarrow$ Sample($g$)\\
  AllocateOnGPU(sampled\_page)\\
  \ForEach{ellpack\_page in X}{
    Compact(sampled\_page, ellpack\_page)\\
  }
  \tcp{Use in-core algorithm}
  tree $\leftarrow$ BuildTree(sampled\_page, $g'$)\\
  \caption{Out-of-Core GPU Tree Construction with Sampling}
  \label{alg:out-of-core-gpu-tree-construction-with-sampling}
\end{algorithm}

To improve the training performance, we implemented gradient-based sampling using MVS.
For each iteration, we first sample the gradient pairs.
Then the multiple ELLPACK pages are compacted together into a single page, only keeping the
rows with non-zero gradients.
Algorithm~\ref{alg:out-of-core-gpu-tree-construction-with-sampling} shows this approach.

\section{Results}
\label{sec:results}
We measured the effectiveness of out-of-core GPU gradient boosting from several dimensions: data
size, model accuracy, and training time.

\subsection{Data Size}
\label{subsec:data-size}
A synthetic dataset with 500 columns is generated using Scikit-learn~\cite{ScikitLearn2011}.
The measurement is done on a Google Cloud Platform (GCP) instance with an NVIDIA Tesla V100 GPU
(16 GiB).
Table~\ref{tab:maximum-data-size} shows the maximum number of rows that can be accommodated in each
mode before getting an out-of-memory error.

\begin{table}
  \caption{Maximum Data Size}
  \label{tab:maximum-data-size}
  \begin{tabular}{lr}
    \toprule
    Mode & \# Rows\\
    \midrule
    In-core GPU & 9 million\\
    Out-of-core GPU & 13 million\\
    Out-of-core GPU, $f=0.1$ & 85 million\\
  \bottomrule
\end{tabular}
\end{table}

Combined with gradient-based sampling, the out-of-core mode allows an order of magnitude bigger
dataset to be trained on a given GPU.
For reference, the 85-million row, 500 column dataset is 903 GiB on disk in LibSVM
format~\cite{Chang2011}, and can be trained successfully on a single 16 GiB GPU using a sampling
ratio of 0.1.

\subsection{Model Accuracy}
\label{subsec:model-accuracy}
\begin{figure*}
  \begin{center}
    \input{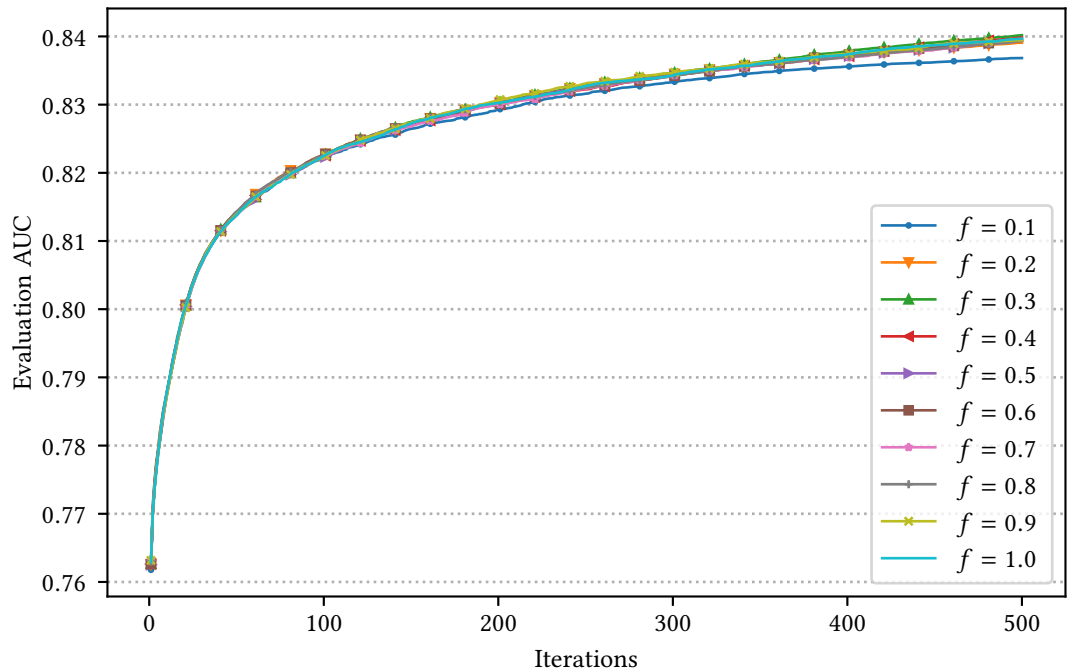}
  \end{center}
  \caption{Training curves on Higgs dataset}
  \label{fig:higgs-training-curves}
\end{figure*}

When not sampling the data, the out-of-core GPU algorithm is equivalent to the in-core version.
With sampling, the size of the data that can fit on a given GPU is increased.
Ideally, this should not change the generalization performance of the trained model.
Figure~\ref{fig:higgs-training-curves} shows the training curves on the Higgs
dataset~\cite{Baldi2014}.
Models with different sampling rates performed similarly, only dropped slightly when $f=0.1$.

For a more detailed evaluation of MVS, see~\cite{Ibragimov2019}.

\subsection{Training Time}
\label{subsec:training-time}
For end-to-end training time, the Higgs dataset is used, split randomly 0.95/0.05 for training and
evaluation.
All the XGBoost parameters use their default value, except that \verb|max_depth| is increased to 8,
and \verb|learning_rate| is lowered to 0.1.
Training is done for 500 iterations.
The hardware used is a desktop computer with an Intel Core i7-5820K processor, 32 GB main memory,
and an NVIDIA Titan V with 12 GiB memory.
Table~\ref{tab:higgs-training-time} shows the training time and evaluation AUC for the different
modes.

Although out-of-core GPU training is slower than the in-core version when sampling is enabled, it
is still significantly faster than the CPU-based algorithm.

\begin{table}
  \caption{Training Time on Higgs Dataset}
  \label{tab:higgs-training-time}
  \begin{tabular}{lrrr}
    \toprule
    Mode & Time(seconds) & AUC\\
    \midrule
    CPU In-core & 1309.64 & 0.8393\\
    CPU Out-of-core & 1228.53 & 0.8393\\
    GPU In-core & 241.52 & 0.8398\\
    GPU Out-of-core, $f=1.0$ & \textbf{211.91} & 0.8396\\
    GPU Out-of-core, $f=0.5$ & 427.41 & 0.8395\\
    GPU Out-of-core, $f=0.3$ & 421.59 & \textbf{0.8399}\\
  \bottomrule
\end{tabular}
\end{table}

\section{Discussion}
\label{sec:discussion}
Faced with the explosive growth of data, GPU proved to be an excellent choice to speed up machine
learning tasks.
However, the relative small size of GPU memory puts a constraint on how much data can be
handled on a single GPU.
To train on larger datasets, distributed algorithms can be used to share the workload on
multiple machines with multiple GPUs.
Setting up and managing a distributed GPU cluster is expensive, both in terms of hardware and
networking cost and system administration overhead.
It is therefore desirable to relax the GPU memory constraint on a single machine, to allow for
easier experimentation with larger datasets.

Because of the PCIe bottleneck, GPU out-of-core computation remains a challenge.
A naive implementation that simply spills data over to main memory or disk would likely to be too
slow to be useful.
If the out-of-core GPU algorithm is slower than the CPU version, then what is the point?
Only by pursuing algorithmic changes, as we have done with gradient-based sampling here, can
out-of-core GPU computation become competitive.
The sampling approach may be applicable to other machine learning algorithms.
This is left as possible future work.

Working with XGBoost also presented unique software engineering challenges.
It is a popular open source project with many contributors, ranging from students, data
scientists, to machine learning software engineers.
Code quality varies between different parts of the code base.
In order to support the existing users, many of which run XGBoost in production, care must be
taken to preserve the current behavior, and plan for breaking changes carefully.
Much of the effort during this project was spent on refactoring the code to make it easier to add
new behaviors.

\section{Conclusion}
\label{sec:conclusion}
In this paper we presented the first ever out-of-core GPU gradient boosting implementation.
This approach greatly expands the size of training data that can fit on a given GPU, without
sacrificing model accuracy or training time.
The source code changes are merged into the open-source XGBoost library.
It is available for production use and further research.

\begin{acks}
We would like to thank Rory Mitchell and Jiaming Yuan for helpful design discussions and careful
code reviews.
Special thanks to Sriram Chandramouli for helping with the implementation, and Philip Hyunsu Cho
for maintaining XGBoost's continuous build system.
\end{acks}

\bibliographystyle{ACM-Reference-Format}
\bibliography{out-of-core-gpu-xgboost}

\end{document}